\documentclass[10pt,twocolumn,letterpaper]{article}

\usepackage{graphicx}
\usepackage{subcaption}
\usepackage{float}
\usepackage[justification=raggedright]{caption}	% makes captions ragged right - thanks to Bryce Lobdell
\usepackage{lscape}                                         % Useful for wide tables or figures.

\usepackage[lined,ruled,linesnumbered]{algorithm2e}

\usepackage{booktabs}                   % Publication quality tables
\usepackage{multirow}

\usepackage{paralist}
\usepackage{enumitem}

\usepackage{bm}                          % Make bold, italic math symbols
\usepackage{epsfig}                      % for figures
\usepackage{graphicx}                  % another package that works for figures
\usepackage{times}
\usepackage{mathptmx}
\usepackage{mathtools}
\usepackage{amssymb,amsmath}   % Short math guide for LaTeX ftp://ftp.ams.org/pub/tex/doc/amsmath/short-math-guide.pdf

\usepackage{units}
\usepackage{color}

\usepackage{comment}

\usepackage{url}  % Hyphenation of URLs.
\usepackage[pagebackref=true,breaklinks=true,letterpaper=true,colorlinks,bookmarks=false,citecolor=blue,linkcolor=blue]{hyperref}
\usepackage{xspace}
\usepackage[table]{xcolor}
\usepackage{setspace}

\def\etal{et~al.\_}			  % and others, and co-workers
\def\eg{e.g.,~}               % for example
\def\ie{i.e.,~}               % that is, in other words
\newlength\paramargin
\newlength\figmargin
\newlength\secmargin
\newlength\figcapmargin

\newcommand{\mpage}[2]
{
\begin{minipage}{#1\linewidth}\centering
#2
\end{minipage}
}

\newcommand{\secref}[1]{Section~\ref{sec:#1}}
\newcommand{\figref}[1]{Figure~\ref{fig:#1}} 
\newcommand{\tabref}[1]{Table~\ref{tab:#1}}
\newcommand{\eqnref}[1]{\eqref{eq:#1}}

\newcommand{\algref}[1]{Algorithm~\ref{alg:#1}}

\long\def\ignorethis#1{}

\def\xi{\mathbf{x}_i}

\graphicspath{{figure}, {images}, {example}}

\makeatletter
\newcommand{\printfnsymbol}[1]{%
  \textsuperscript{\@fnsymbol{#1}}%
}
\makeatother

\usepackage{cvpr}

\cvprfinalcopy % *** Uncomment this line for the final submission

\def\cvprPaperID{6851} % *** Enter the CVPR Paper ID here

\ifcvprfinal\fi

\begin{document}

%%%%%%%%% TITLE
%\title{DropGrad: Gradient Dropout Regularization for Meta-Learning}
\title{Regularizing Meta-Learning via Gradient Dropout}
\author{{Hung-Yu Tseng\thanks{Equal contribution}\hspace{5pt}}$^1$, Yi-Wen Chen\printfnsymbol{1}$^1$, Yi-Hsuan Tsai$^2$, Sifei Liu$^3$, Yen-Yu Lin$^4$, Ming-Hsuan Yang$^{1,5}$ \vspace{3.5mm}\\
$^{1}$University of California, Merced\hspace{20pt}$^{2}$NEC Laboratories America\\
$^{3}$Nvidia Research\hspace{20pt}$^{4}$National Chiao Tung University\hspace{20pt}$^{5}$Google Research}

\maketitle

\begin{abstract}
\vspace{-2mm}

With the growing attention on learning-to-learn new tasks using only a few examples,
meta-learning has been widely used in numerous problems such as few-shot classification, reinforcement learning, and domain generalization.
However, meta-learning models are prone to overfitting when there are no sufficient training tasks for the meta-learners to generalize. 
Although existing approaches such as Dropout are widely used to address the overfitting problem, these methods are typically designed for regularizing models of a single task in supervised training.
In this paper, we introduce a simple yet effective method to alleviate the risk of overfitting for gradient-based meta-learning.
Specifically, during the gradient-based adaptation stage, we randomly drop the gradient in the inner-loop optimization of each parameter in deep neural networks, such that the augmented gradients improve generalization to new tasks.
We present a general form of the proposed gradient dropout regularization and show that this term can be sampled from either the Bernoulli or Gaussian distribution.
To validate the proposed method, we conduct extensive experiments and analysis on numerous computer vision tasks, demonstrating that the gradient dropout regularization mitigates the overfitting problem and improves the performance upon various gradient-based meta-learning frameworks.
\end{abstract}
\vspace{-4mm}

\section{Introduction}
\label{sec:intro}

In recent years, significant progress has been made in meta-learning, which is also known as \textit{learning to learn}.
One common setting is that, given only a few training examples, meta-learning aims to learn new \textit{tasks} rapidly by leveraging the past experience acquired from the known tasks.
It is a vital machine learning problem due to the potential for reducing the amount of data and time for adapting an existing system.
Numerous recent methods successfully demonstrate how to adopt meta-learning algorithms to solve various learning problems, such as few-shot classification~\cite{finn2017maml,santoro2016meta,snell2017prototypical}, reinforcement learning~\cite{gupta2018meta,rakelly2019efficient}, and domain generalization~\cite{balaji2018metareg,li2018learning}.

Despite the demonstrated success, meta-learning frameworks are prone to overfitting~\cite{kim2018bmaml} when there do not exist sufficient training tasks for the meta-learners to generalize.
For instance, few-shot classification on the mini-ImageNet~\cite{vinyals2016matching} dataset contains only $64$ training categories.
Since the training tasks can be only sampled from this small set of classes, meta-learning models may overfit and fail to generalize to new testing tasks.
%Since the training tasks can be only sampled from a small dataset, meta-learning models may overfit and fail to generalize to new testing tasks, as illustrated in \figref{validatecurve}.
%of the experimental section.

Significant efforts have been made to address the overfitting issue in the supervised learning framework, where the model is developed to learn a \textit{single} task (\eg, recognizing the same set of categories in both training and testing phases).
The Dropout~\cite{srivastava2014dropout} method randomly drops~(zeros) intermediate activations in deep neural networks during the training stage.
Relaxing the limitation of binary dropout, the Gaussian dropout~\cite{wang2013fast} scheme augments activations with noise sampled from a Gaussian distribution.
Numerous methods~\cite{ghiasi2018dropblock,larsson2016droppath,tompson2015spatialdropout,wan2013dropconnect,zoph2018scheduledroppath} further improve the Dropout method by injecting structural noise or scheduling the dropout process to facilitate the training procedure.
Nevertheless, these methods are developed to regularize the models to learn a single task, which may not be effective for meta-learning frameworks.

In this paper, we address the overfitting issue~\cite{kim2018bmaml} in gradient-based meta-learning.
As shown in~\figref{illustration}(a), given a new task, the meta-learning framework aims to adapt model parameters $\theta$ to be $\theta'$ via the gradients computed according to the few examples (support data $\mathcal{X}^s$).
This gradient-based adaptation process is also known as the \emph{inner-loop} optimization.
To alleviate the overfitting issue, one straightforward approach is to apply the existing dropout method to the model weights directly.
However, there are two sets of model parameters $\theta$ and $\theta'$ in the inner-loop optimization.
As such, during the meta-training stage, applying normal dropout would cause inconsistent randomness, \ie, dropped neurons, between these two sets of model parameters.
To tackle this issue, we propose a dropout method on the gradients in the inner-loop optimization, denoted as \textit{DropGrad}, to regularize the training procedure.
This approach naturally bridges $\theta$ and $\theta'$, and thereby involves only one randomness for the dropout regularization.
%In this paper, we address the overfitting issue~\cite{kim2018bmaml} in gradient-based meta-learning for adapting a model to new tasks.
%
%One straightforward scheme is to apply the existing dropout method to the model weights directly.
%
%However, in gradient-based meta-learning frameworks, there are two sets of model parameters, \ie, $\theta$ and $\theta'$ for the support set $\mathbf{x}^s$ and the query set $\mathbf{x}^q$, respectively (see Figure~\ref{fig:illustration}(a)).
%
%As such, during the meta-training stage, applying normal dropout would cause inconsistent updates due to randomnesses on $\theta$ and $\theta'$.
%
%To tackle this issue, we instead propose a dropout method on gradients, denoted as \textit{DropGrad}, to regularize the training procedure which naturally bridges $\theta$ and $\theta'$, and thereby involves only the same randomness on gradients for the dropout regularization.
%
%More specifically, during the meta-training stage, the objective is to minimize the loss of the new task on $\mathbf{x}^q$ and adapt the model parameters $\theta'$. Then, given the gradients that are obtained from $\mathbf{x}^s$ in the inner-loop optimization, our DropGrad method further augments these gradients with a noise sampling process (see Figure~\ref{fig:illustration}(a)).
%
We also note that our method is model-agnostic and generalized to various gradient-based meta-learning frameworks such as~\cite{antoniou2018maml++,finn2017maml,li2017metasgd}.
In addition, we demonstrate that the proposed dropout term can be formulated in a general form, where either the binary or Gaussian distribution can be utilized to sample the noise, as demonstrated in~\figref{illustration}(b).
% and can be related in terms of a given dropout probability.
% - where we augment the gradients and when we update them in the optimization step, can also refer to Figure 1 and mention the differences \\
% - we formulate our dropout operation in a form that binary and Gaussian dropouts can be both utilized and compared

To evaluate the proposed DropGrad method, we conduct experiments on numerous computer vision tasks, including few-shot classification on the mini-ImageNet~\cite{vinyals2016matching}, online object tracking~\cite{park2018meta}, and few-shot viewpoint estimation~\cite{tseng2019metaview}, showing that the DropGrad scheme can be applied to and improve different tasks.
In addition, we present comprehensive analysis by using various meta-learning frameworks, adopting different dropout probabilities, and explaining which layers to apply gradient dropout.
% \sliu{what is gradient layer?}.
%
To further demonstrate the generalization ability of DropGrad, we perform a challenging cross-domain few-shot classification task, in which the meta-training and meta-testing sets are from two different distributions, \ie, the mini-ImageNet and CUB~\cite{WelinderEtal2010cub} datasets.
We show that with the proposed method, the performance is significantly improved under the cross-domain setting.
We make the source code public available to simulate future research in this field.\footnote{\url{https://github.com/hytseng0509/DropGrad}}
% the overfitting issue in the cross-domain setting is mitigated and thus the performance is significantly improved.
%
%\textcolor{blue}{
%To visualize the image regions where CNN models focus on when performing classification, we generate the class activation maps (CAMs)~\cite{zhou2016learning} under the cross-domain setting.
%
%We show that models trained with the proposed DropGrad method are able to focus more on the objects, which are the discriminative regions in images.
%}

% - overview of the experimental setting \\
% - what are the applications/tasks we conduct experiments \\
% - what are the ablation study and analysis we provide

%\paragraph{List of contributions}
% Create a list of contributions. This helps the reviewers to summarize the paper. The rest of the paper provide evidences for the claimed novelty/contributions. Use forward references to provide a roadmap to the paper, e.g., we propose XXX (\secref{algorithm})
In this paper, we make the following contributions:
%\begin{itemize}
\begin{compactitem}
\item We propose a simple yet effective gradient dropout approach to improve the generalization ability of gradient-based meta-learning frameworks.
\item We present a general form for gradient dropout and show that both binary and Gaussian sampling schemes mitigate the overfitting issue.
\item We demonstrate the effectiveness and generalizability of the proposed method via extensive experiments on numerous computer vision tasks.
\end{compactitem}
%\end{itemize}

\section{Related Work}

\vspace{-2mm}
\paragraph{Meta-Learning.}
Meta-learning aims to adapt the past knowledge learned from previous tasks to new tasks with few training instances.
Most meta-learning algorithms can be categorized into three groups:
1) Memory-based approaches~\cite{rezende2016oneshot,santoro2016meta} utilize recurrent networks to process few training examples of new tasks sequentially;
2) Metric-based frameworks~\cite{oreshkin2018tadam,snell2017prototypical,sung2018learning,vinyals2016matching,tseng2020cross} make predictions by referring to the features encoded from the input data and training instances in a generic metric space;
3) Gradient-based methods~\cite{antoniou2018maml++,finn2017maml,finn2018pmaml,kim2018bmaml,li2017metasgd,rusu2018leo,ravi2018amortized} learn to optimize the model via gradient descent with few examples, which is the focus of this work.  
In the third group, the MAML~\cite{finn2017maml} approach learns model initialization (\ie, initial parameters) that is amenable to fast fine-tuning with few instances.
In addition to model initialization, the MetaSGD~\cite{li2017metasgd} method learns a set of learning rates for different model parameters.
Furthermore, the MAML++~\cite{antoniou2018maml++} algorithm makes several improvements based on the MAML method to facilitate the training process with additional performance gain.
However, these methods are still prone to overfitting as the dataset for the training tasks is insufficient for the model to adapt well. 
Recently, Kim~\etal~\cite{kim2018bmaml} and Rusu~\etal~\cite{rusu2018leo} address this issue via the Bayesian approach and latent embeddings. 
Nevertheless, these methods employ additional parameters or networks which entail significant computational overhead and may not be applicable to arbitrary frameworks.
In contrast, the proposed regularization does not impose any overhead and thus can be readily integrated into the gradient-based models mentioned above.

\vspace{-2mm}
\paragraph{Dropout Regularization.}
Built upon the Dropout~\cite{srivastava2014dropout} method, various schemes~\cite{ghiasi2018dropblock,goodfellow2013maxout,larsson2016droppath,tompson2015spatialdropout,wan2013dropconnect} have been proposed to regularize the training process of deep neural networks for supervised learning.
The core idea is to inject noise into intermediate activations when training deep neural networks.
Several recent studies improve the regularization on convolutional neural networks by making the injected structural noise.
For instance, the SpatialDropout~\cite{tompson2015spatialdropout} method drops the entire channel from an activation map, the DropPath~\cite{larsson2016droppath,zoph2018scheduledroppath} scheme chooses to discard an entire layer, and the DropBlock~\cite{ghiasi2018dropblock} algorithm zeros multiple continuous regions in an activation map.
%
%MH: I am not sure what you mean by single task, and then your method is for "task-level". Both are on tasks".  Do you mean multiple tasks?
%HY: revise and add an example
Nevertheless, these approaches are designed for deep neural networks that aim to learn a \textit{single} task, \eg, learning to recognize a fixed set of categories.
In contrast, our algorithm aims to regularize the gradient-based meta-learning frameworks that suffer from the overfitting issue on the \textit{task}-level, \eg, introducing new tasks.
%On the other hand, our algorithm aims to regularize the gradient-based meta-learning frameworks which suffer from the overfitting issue on the \textit{task}-level, \eg, introducing new tasks.
\section{Gradient Dropout Regularization}

\begin{figure*}[t]
    \centering
    \includegraphics[width=0.9\textwidth]{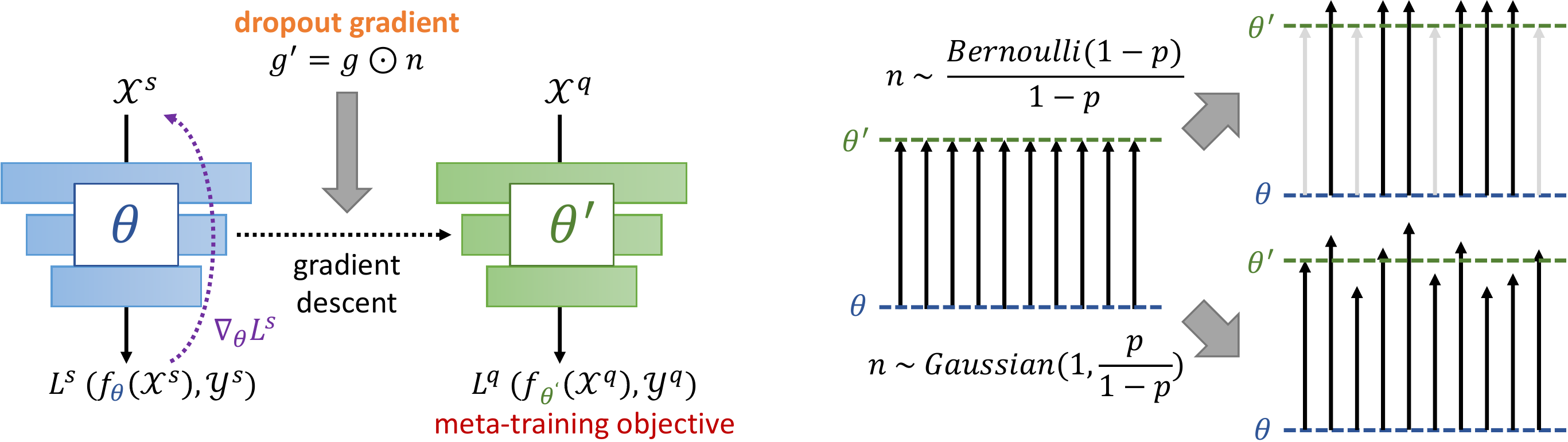}\\
    \mpage{0.49}{(a)} \hfill
    \mpage{0.49}{(b)}
    \vspace{-4mm}
    \caption{
    \textbf{Illustration of the proposed method.} 
    (a) The proposed DropGrad method imposes a noise term $n$ to augment the gradient in the inner-loop optimization during the meta-training stage.
    (b) The DropGrad method samples the noise term $n$ from either the Bernoulli or Gaussian distribution, in which the Gaussian distribution provides a better way to account for uncertainty.
    }
    \label{fig:illustration}
    \vspace{-4mm}
\end{figure*}

Before introducing details of our proposed dropout regularization on gradients, we first review the gradient-based meta-learning framework.

%MH: we can also move this Section 3.3 as Discussions after introducing the method. It is best to add more discussions to show that you understand this method well (when and why it works well).
%HY: move this part to section 3.2
%The Dropout~\cite{srivastava2014dropout} method introduces noise to the neural networks during training to reduce the overfitting problem. 
%
%However, introducing noise in gradient-based meta-learning may not be effective due to the inconsistency between two different sets of parameters q and q0 in the single forward pass. 
%
%To address this issue, we perform the dropout process
% on gradients which bridge the two sets of parameters. 
%
%Note that the proposed DropGrad scheme is also related to the simulated annealing (SA)~\cite{kirkpatrick1983optimization} method.  
%
%While the SA method modulates gradients by exploring uncertain solutions to escape from a local minimum, the DropGrad method, drops the inner gradient to
%introduce uncertainty in the forward pass of the gradient-based meta-learning framework. 
%
%While the SA method and the proposed DropGrad are conceptually similar, their goals and formulations are significantly different.

\subsection{Preliminaries for Meta-Learning}
\label{sec:3_1}
In meta-learning, multiple tasks $\mathcal{T}=\{T_1, T_2, ..., T_n\}$ are divided into meta-training $\mathcal{T}^\mathrm{train}$, meta-validation $\mathcal{T}^\mathrm{val}$, and meta-testing $\mathcal{T}^\mathrm{test}$ sets.
Each task $T_i$ consists of a support set $D^{s}={(\mathcal{X}^s, \mathcal{Y}^s)}$ and a query set $D^{q}={(\mathcal{X}^q, \mathcal{Y}^q)}$, where $\mathcal{X}$ and $\mathcal{Y}$ are a set of input data and the corresponding ground-truth.
The support set $D^{s}$ represents the set of few labeled data for learning, while the query set $D^{q}$ indicates the set of data to be predicted.

Given a novel task and a parametric model $f_\theta$, the objective of a gradient-based approach during the meta-training stage is to minimize the prediction loss $L^q$ on the query set $D^q$ according to the signals provided from the support set $D^s$, and thus the model $f_\theta$ can be adapted.
% after the model $f_\theta$ is adapted according to the signal given in the support set $D^s$.
%the goal of MAML is to learn a set of model initial parameters $\theta$ that is amendable to successful adaptation to the new task via gradient descent.
%
%Here we give an overview of the MAML~\cite{finn2017maml} method, which offers a general formulation of gradient-based frameworks.
\figref{illustration}(a) shows an overview of the MAML~\cite{finn2017maml} method, which offers a general formulation of gradient-based frameworks.
For each iteration of the meta-training phase, we first randomly sample a task $T=\{D^{s}, D^{q}\}$ from the meta-training set $\mathcal{T}^\mathrm{train}$.
We then adapt the initial parameters $\theta$ to be task-specific parameters $\theta'$ via gradient descent:
% \sliu{no ``$\times$''?}:
\begin{equation}
\theta' = \theta - \alpha \odot {g},
\label{eq:eq1}
\end{equation}
where $\alpha$ is the learning rate for gradient-based adaptation and $\odot$ is the operation of element-wise product, \ie, Hadamard product.
The term $g$ in \eqref{eq:eq1} is the set of gradients computed according to the objectives of model $f_\theta$ on the support set $D^{s}=(\mathcal{X}^s, \mathcal{Y}^s)$:
\begin{equation}
g = \bigtriangledown_{\theta}L^s(f_\theta(\mathcal{X}^s), \mathcal{Y}^s).
\label{eq:gradient}
\end{equation}
We call the step of \eqref{eq:eq1} as the inner-loop optimization and
typically, we can do multiple gradient steps for \eqref{eq:eq1}, \eg, smaller than $10$ in general.
After the gradient-based adaptation, the initial parameters $\theta$ are optimized according to the loss functions of the adapted model $f_{\theta'}$ on the query set $D^{q}=(\mathcal{X}^q, \mathcal{Y}^q)$:
\begin{equation}
\theta = \theta - \eta\bigtriangledown_{\theta}L^q(f_{\theta'}(\mathcal{X}^q), \mathcal{Y}^q),
\end{equation}
where $\eta$ is the learning rate for meta-training.
During the meta-testing stage, the model $f_\theta$ is adapted according to the support set $D^{s}$ and the prediction on query data $\mathcal{X}^q$ is made without accessing the ground-truth $\mathcal{Y}^q$ in the query set.
%
% We show the overview of a gradient-based framework in \figref{illustration}(a).
%
We note that several methods are built upon the above formulation introduced in the MAML method. 
For example, the learning rate $\alpha$ for gradient-adaptation is viewed as the optimization objective~\cite{antoniou2018maml++,li2017metasgd}, and the initial parameters $\theta$ are not generic but conditional on the support set $D^{s}$~\cite{rusu2018leo}.

\subsection{Gradient Dropout}

%The Dropout~\cite{srivastava2014dropout} method introduces noise to the neural networks during training to reduce the overfitting problem. 
%
%However, introducing noise in gradient-based meta-learning may not be effective due to the inconsistency between two different sets of parameters q and q0 in the single forward pass. 
%
%To address this issue, we perform the dropout process
%on gradients which bridge the two sets of parameters. 
%
%Note that the proposed DropGrad scheme is also related to the simulated annealing (SA)~\cite{kirkpatrick1983optimization} method.  
%
%While the SA method modulates gradients by exploring uncertain solutions to escape from a local minimum, the DropGrad method, drops the inner gradient to
%introduce uncertainty in the forward pass of the gradient-based meta-learning framework. 
%
%While the SA method and the proposed DropGrad are conceptually similar, their goals and formulations are significantly different.
%
%\textcolor{red}{
The main idea is to impose uncertainty to the core objective during the meta-training step, \ie, the gradient $g$ in the inner-loop optimization, such that $\theta'$ receives gradients with noise to improve the generalization of gradient-based models.
% receive a more diverse signals that prevent the model from overfitting.
%}
%The main idea is to impose uncertainty to the core objective during the meta-training step, \ie, adapting the model $\theta'$, such that $\theta'$ can receive a more diverse signals that prevent the model from overfitting.
% in the forward pass of a model.
%
As described in Section~\ref{sec:3_1}, adapting the model $\theta$ to $\theta'$ involves the gradient update in the inner-loop optimization formulated in \eqref{eq:gradient}.
% the forward pass of a gradient-based meta-learning model involves the gradient adaption described in \eqref{eq:eq1}.
%
Based on this observation, we propose to randomly drop the gradient in \eqref{eq:gradient}, \ie, $g$, during the inner-loop optimization, as illustrated in \figref{illustration}.
Specifically, we augment the gradient g as follows:
\begin{equation}
g' = g \odot {n},
\label{eq:noise}
\end{equation}
where $n$ is a noise regularization term sampled from a pre-defined distribution.
With the formulation of \eqref{eq:noise}, in the following we introduce two noise regularization strategies via sampling from different distributions, \ie, the Bernoulli and Gaussian distributions.

\vspace{-2mm}
\paragraph{Binary DropGrad.}
We randomly zero the gradient with the probability $p$,
in which the process can be formulated as:
\begin{equation}
g' = g\odot{n_b},\hspace{3mm}n_b \sim \frac{Bernoulli(1 - p)}{1 - p},
\label{eq:binary}
\end{equation}
where the denominator $1 - p$ is the normalization factor.
Note that, different from the Dropout~\cite{srivastava2014dropout} method which randomly drops the intermediate activations in a supervised learning network under a single task setting, we perform the dropout on the gradient level.

\vspace{-2mm}
\paragraph{Gaussian DropGrad.}
One limitation of the Binary DropGrad scheme is that the noise term $n_b$ is only applied in a binary form, which is either $0$ or $1-p$.
To address this disadvantage and provide a better regularization with uncertainty, we extend the Bernoulli distribution to the Gaussian formulation.
% \sliu{you mean you draw links between them?}
%
Since the expectation and variance of the noise term $n_b$ in the Binary DropGrad method are respectively $\mathrm{E}(n_b)=1$ and $\sigma^2(n_b)=\frac{p}{1-p}$, we can augment the gradient $g$ with noise sampled from the Gaussian distribution:
\begin{equation}
g' = g\odot{n_g},\hspace{3mm}n_g \sim Gaussian(1, \frac{p}{1-p}).
\label{eq:gaussian}
\end{equation}
As a result, two noise terms $n_b$ and $n_g$ are statistically comparable with the same dropout probability $p$.
% \sliu{how?}
%
In \figref{illustration}(b), we illustrate the difference between the Binary DropGrad and Gaussian DropGrad approaches.
We also show the process of applying the proposed regularization using the MAML~\cite{finn2017maml} method in \algref{alg}, while similar procedures can be applied to other gradient-based meta-learning frameworks, such as MetaSGD~\cite{li2017metasgd} and MAML++~\cite{antoniou2018maml++}.

\begin{algorithm}[t]
  \caption{Applying DropGrad on MAML~\cite{finn2017maml}}
  \label{alg:alg}
  \DontPrintSemicolon
  \textbf{Require:} a set of training tasks $\mathcal{T}^\mathrm{train}$, adaptation learning rate $\alpha$, meta-learning rate $\eta$\;
  randomly initialize $\theta$\;
  \While{training} {
    randomly sample a task $T=\{D^s(\mathcal{X}^s, \mathcal{Y}^s), D^q(\mathcal{X}^q, \mathcal{Y}^q)\}$ from $\mathcal{T}^\mathrm{train}$\;

    $g = \bigtriangledown_{\theta}L^s(f_\theta(\mathcal{X}^s), \mathcal{Y}^s)$\;
    
    compute $g'$ according to \eqnref{binary} or \eqnref{gaussian}\;
    $\theta' = \theta - \alpha\times{g'}$\;
    $\theta = \theta - \eta\bigtriangledown_{\theta}L^q(f_{\theta'}(\mathcal{X}^q), \mathcal{Y}^q)$\;
  }
\end{algorithm}

\section{Experimental Results}
\label{sec:exp}
In this section, we evaluate the effectiveness of the proposed DropGrad method by conducting extensive experiments on three learning problems: few-shot classification, online object tracking, and few-shot viewpoint estimation.
In addition, for the few-shot classification experiments, we analyze the effect of using binary and Gaussian noise, which layers to apply DropGrad, and performance in the cross-domain setting.
%
%The source code and trained models will be made available to the public.

%\vspace{-2mm}
\subsection{Few-Shot Classification}
%\vspace{-2mm}

Few-shot classification aims to recognize a set of new categories, \eg, five categories~($5$-way classification), with few, \eg, one~($1$-shot) or five~($5$-shot), example images from each category.
In this setting, the support set $D^s$ contains the few images $\mathcal{X}^s$ of the new categories and the corresponding categorical annotation $\mathcal{Y}^s$.
We conduct experiments on the mini-ImageNet~\cite{vinyals2016matching} dataset, which is widely used for evaluating few-shot classification approaches.
As a subset of the ImageNet~\cite{deng2009imagenet}, the mini-ImageNet dataset contains $100$ categories and $600$ images for each category.
We use the $5$-way evaluation protocol in~\cite{ravi2017metalstm} and split the dataset into $64$ training, $16$ validating, and $20$ testing categories.

\vspace{-2mm}
\paragraph{Implementation Details.}
We apply the proposed DropGrad regularization method to train the following gradient-based meta-learning frameworks: MAML~\cite{finn2017maml}, MetaSGD~\cite{li2017metasgd}, and MAML++~\cite{antoniou2018maml++}.
%
%MH: what do you mean by "upon the MAML one"?
We use the implementation from Chen~\etal~\cite{chen2019closerfewshot} for MAML and use our own implementation for MetaSGD.\footnote{https://github.com/wyharveychen/CloserLookFewShot}
We use the ResNet-18~\cite{he2016deep} model as the backbone network for both MAML and MetaSGD.
As for MAML++, we use the original source code.\footnote{https://github.com/AntreasAntoniou/HowToTrainYourMAMLPytorch}
Similar to recent studies~\cite{rusu2018leo}, we also pre-train the feature extractor of ResNet-18 by minimizing the classification loss on the $64$ training categories from the mini-ImageNet dataset for the MetaSGD method, which is denoted by MetaSGD*.

For all the experiments, we use the \emph{default hyper-parameter settings} provided by the original implementation.
Moreover, we select the model according to the validation performance for evaluation~(\ie, \emph{early stopping}).

% ------------------------------------------------------------ %
\begin{figure}[t]
    \centering
    \includegraphics[width=\linewidth]{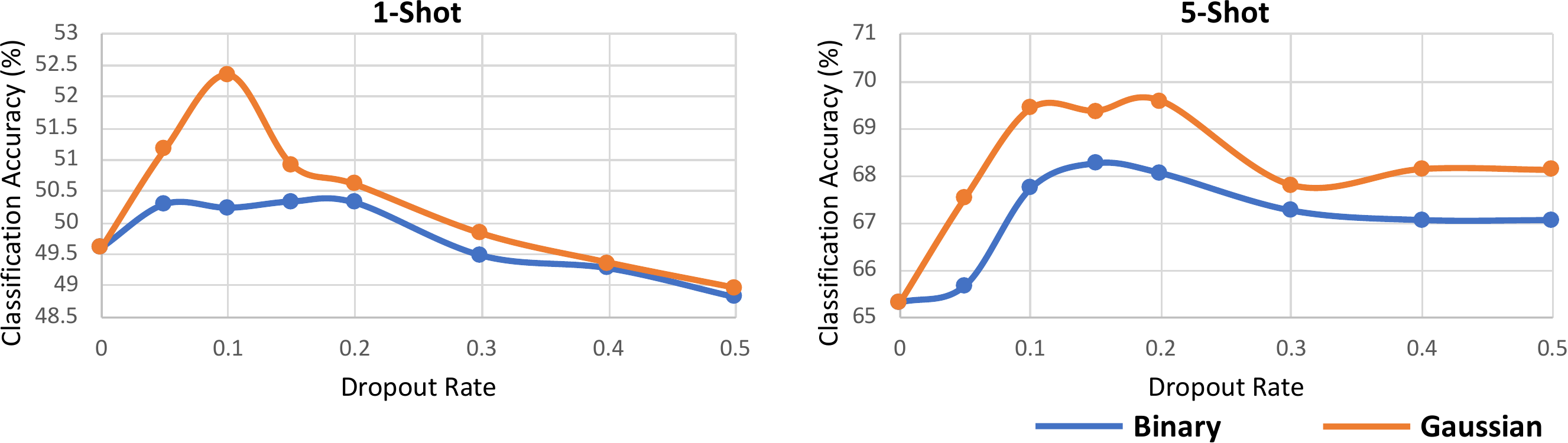}
    %\vspace{-2mm}
    \caption{\textbf{Comparison between the proposed Binary and Gaussian DropGrad methods.} We compare the $1$-shot (\textit{left}) and $5$-shot (\textit{right}) performance of MAML~\cite{finn2017maml} trained with two different forms of DropGrad under various dropout rates on mini-ImageNet. 
    %The proposed DropGrad method is particularly effective with the dropout rate in $[0.1, 0.2]$. Moreover, the Gaussian DropGrad method consistently obtains better results compared to the Binary DropGrad scheme. 
    %Therefore, we apply the Gaussian DropGrad method with the dropout rate of $0.1$ or $0.2$ in all of our experiments.
    }
    \label{fig:binarygaussian}
    \vspace{-4mm}
\end{figure}
% ------------------------------------------------------------ %

\vspace{-2mm}
\paragraph{Comparison between Binary and Gaussian DropGrad.}
We first evaluate how the proposed Binary and Gaussian DropGrad methods perform on the MAML framework with different values of the dropout probability $p$.
\figref{binarygaussian} shows that both methods are effective especially when the dropout rate is in the range of $[0.1, 0.2]$, while setting the dropout rate to $0$ is to turn the proposed DropGrad method off.
Since the problem of learning from only one instance ($1$-shot) is more complicated, the overfitting effect is less severe compared to the $5$-shot setting.
As a result, applying the DropGrad method with a dropout rate larger than $0.3$ degrades the performance.
Moreover, the Gaussian DropGrad method consistently outperforms the binary case on both $1$-shot and $5$-shot tasks, due to a better regularization term $n_g$ with uncertainty.
We then apply the Gaussian DropGrad method with the dropout rate of $0.1$ or $0.2$ in the following experiments.

\vspace{-2mm}
\paragraph{Comparison with Existing Dropout Methods.}
To show that the proposed DropGrad method is effective for gradient-based meta-learning frameworks, we compare it with two existing dropout schemes applied on the network activations in both $f_\theta$ and $f_\theta'$.
We choose the Dropout~\cite{srivastava2014dropout} and SpatialDropout~\cite{tompson2015spatialdropout} methods, since the former is a commonly-used approach while the latter is shown to be effective for applying to $2$D convolutional maps.
The performance of MAML on $5$-shot classification on the mini-ImageNet dataset is: \textit{DropGrad} $69.42 \pm 0.73\%$, \textit{SpatialDropout} $68.09 \pm 0.56\%$, and \textit{Vanilla Dropout} $67.44 \pm 0.57\%$.
This demonstrates the benefit of using the proposed DropGrad method, which effectively tackles the issue of inconsistent randomness between two different models $f_\theta$ and $f_\theta'$ in the inner-loop optimization of gradient-based meta-learning frameworks.

% ------------------------------------------------------------ %
\begin{table}[t]
	\centering
	\footnotesize
	\setlength\tabcolsep{5pt}
	\caption{\textbf{Few-shot classification results on mini-ImageNet.} The Gaussian DropGrad method improves the performance of gradient-based models on $1$-shot and $5$-shot classification tasks.}
	\begin{tabular}{l cc}
	    \toprule
		Model & $1$-shot & $5$-shot \\
		\midrule
		MAML~\cite{finn2017maml} & $49.61 \pm 0.92\%$ & $65.72 \pm 0.77\%$ \\
		MAML w/ Gaussian DropGrad & $\mathbf{52.35 \pm 0.86\%}$ & $\mathbf{69.42 \pm 0.73\%}$ \\
		\midrule
		MetaSGD~\cite{li2017metasgd} & $51.51 \pm 0.87\%$ & $69.67 \pm 0.75\%$ \\
		MetaSGD w/ Gaussian DropGrad & $\mathbf{53.38 \pm 0.93\%}$ & $\mathbf{71.14 \pm 0.72\%}$ \\
		\midrule
		MetaSGD* & $60.44 \pm 0.87\%$ & $72.55 \pm 0.54\%$ \\
		MetaSGD* w/ Gaussian DropGrad & $\mathbf{61.69 \pm 0.84\%}$ & $\mathbf{73.33 \pm 0.57\%}$ \\
		\midrule
    	MAML++~\cite{antoniou2018maml++} & $50.21 \pm 0.50\%$ & $68.66 \pm 0.46\%$ \\
		MAML++ w/ Gaussian DropGrad & $\mathbf{51.13 \pm 0.50\%}$ & $\mathbf{69.80 \pm 0.46\%}$ \\
		\bottomrule 
	\end{tabular}
	\label{tab:mini}
	\vspace{-2mm}
\end{table}
\begin{figure}[t]
    \centering
    \includegraphics[width=\linewidth]{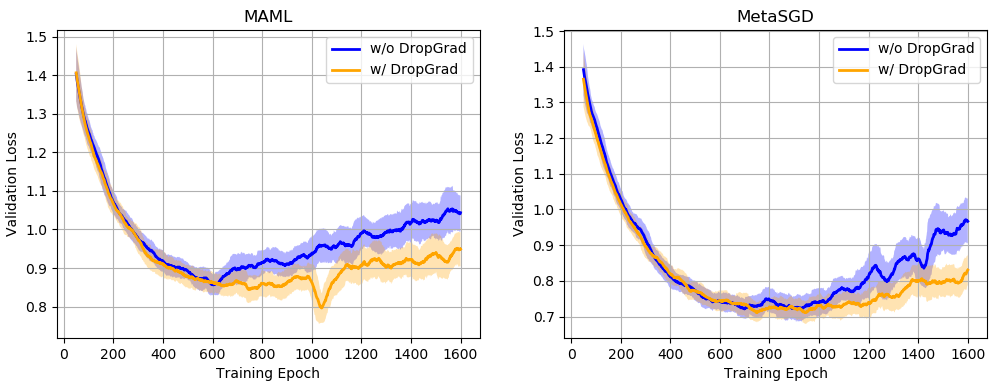}
    \caption{\textbf{Validation loss over training epochs.} We show the validation curves of the MAML~(\textit{left}) and MetaSGD~(\textit{right}) frameworks trained on the $5$-shot mini-ImageNet dataset. 
    %The curves and shaded regions represent the mean and standard deviation of validation loss over 50 epochs. 
    %The curves validate that the proposed DropGrad method alleviates the overfitting problem.
    }
    \label{fig:validatecurve}
    \vspace{-4mm}
\end{figure}
% ------------------------------------------------------------ %

\vspace{-2mm}
\paragraph{Overall Performance on the Mini-ImageNet Dataset.}
\tabref{mini} shows the results of applying the proposed Gaussian DropGrad method to different frameworks.
The results validate that the proposed regularization scheme consistently improves the performance of various gradient-based meta-learning approaches.
In addition, we present the curve of validation loss over training episodes from MAML and MetaSGD on the $5$-shot classification task in \figref{validatecurve}.
We observe that the overfitting problem is more severe in training the MetaSGD method since it consists of more parameters to be optimized.
The DropGrad regularization method mitigates the overfitting issue and facilitates the training procedure.

% ------------------------------------------------------------ %
\begin{table*}[t]
	\centering
	\footnotesize
	\caption{\textbf{Performance of applying DropGrad to different layers.} We conduct experiments on the $5$-shot classification task using MAML on mini-ImageNet. It is more helpful to drop the gradients closer to the output layers (\eg, FC and Block$4$ + FC).}
	\begin{tabular}{c|cccccc} 
	    \toprule
		Origin & FC & Block$4$ + FC & Full & Block$1$ + Conv & Conv \\
		\midrule
		$65.72 \pm 0.77\%$ & $68.93 \pm 0.55\%$ & $69.02 \pm 0.57\%$ & $69.42 \pm 0.73\%$ & $64.96 \pm 0.80\%$ & $65.53 \pm 0.75\%$\\
		\bottomrule 
	\end{tabular}
	\label{tab:where}
	\vspace{-2mm}
\end{table*}
% ------------------------------------------------------------ %

\vspace{-2mm}
\paragraph{Layers to Apply DropGrad.}
We study which layers in the network to apply the DropGrad regularization in this experiment.
The backbone ResNet-18 model contains a convolutional layer~(Conv) followed by $4$ residual blocks~(Block$1$, Block$2$, Block$3$, Block$4$) and a fully-connected layer~(FC) as the classifier.
We perform the Gaussian DropGrad method on different parts of the ResNet-18 model for MAML on the $5$-shot classification task.
The results are presented in \tabref{where}.
We find that it is more critical to drop the gradients closer to the output layers~(\eg, FC and Block$4$ + FC).
Applying the DropGrad method to the input side~(\eg, Block$1$ + Conv and Conv), however, may even negatively affect the training and degrade the performance.
This can be explained by the fact that features closer to the output side are more abstract and thus tend to overfit.
As using the DropGrad regularization term only increases a negligible overhead, we use the \textit{Full} model, where our method is applied to all layers in the experiments unless otherwise mentioned.

% ------------------------------------------------------------ %
\begin{table*}[t]
	\centering
	\footnotesize
	\caption{\textbf{5-shot classification results of MAML under various hyper-parameter settings.} We study the learning rate $\alpha$ and number of iterations $n_\mathrm{inner}$ in the inner-loop optimization of MAML using mini-ImageNet dataset.}
	\begin{tabular}{l cccccc} 
	    \toprule
		 $\alpha$, $n_\mathrm{inner}$ & $0.01$, $5$ (original) & $0.1$, $5$ & $0.001$, $5$ & $0.01$, $3$ & $0.01$, $7$ \\
		\midrule
		MAML~\cite{finn2017maml} &
		$65.72 \pm 0.77\%$ & $65.98 \pm 0.79\%$ & $58.55 \pm 0.80\%$ & $64.84 \pm 0.80\%$ & $68.11 \pm 0.74\%$\\
		MAML w/ DropGrad &
		$\mathbf{69.42 \pm 0.73\%}$ & $\mathbf{67.78 \pm 0.73\%}$ & $\mathbf{64.05 \pm 0.79\%}$ & $\mathbf{65.42 \pm 0.80\%}$ & $\mathbf{69.65 \pm 0.70\%}$\\
		\bottomrule 
	\end{tabular}
	\label{tab:hyperparameter}
	\vspace{-4mm}
\end{table*}
% ------------------------------------------------------------ %

\vspace{-2mm}
\paragraph{Hyper-Parameter Analysis.}
In all experiments shown in \secref{exp}, we use the default hyper-parameter values from the original implementation of the adopted methods.
In this experiment, we explore the hyper-parameter choices for MAML~\cite{finn2017maml}.
Specifically, we conduct an ablation study on the learning rate $\alpha$ and the number of inner-loop optimizations $n_\mathrm{inner}$ in MAML.
As shown in \tabref{hyperparameter}, the proposed DropGrad method improves the performance consistently under different sets of hyper-parameters. 
%

% ------------------------------------------------------------ %
\begin{table}[t]
	\centering
	%\scriptsize
	\footnotesize
	\caption{\textbf{Cross-domain performance for few-shot classification.} We use the mini-ImageNet and CUB datasets for the meta-training and meta-testing steps, respectively.
	The improvement of applying the proposed DropGrad method is more significant in the cross-domain cases than the intra-domain ones.}
	\begin{tabular}{l cc} 
	    \toprule
		Model & $1$-Shot & $5$-Shot \\
		\midrule
		MAML~\cite{finn2017maml} & 
		$31.52 \pm 0.52\%$ & $45.56 \pm 0.51\%$ \\
		MAML w/ Dropout~\cite{srivastava2014dropout} &
		$31.84 \pm 0.49\%$ & $46.48 \pm 0.50\%$ \\
		MAML w/ DropGrad & 
		$\mathbf{33.20 \pm 0.67\%}$ & $\mathbf{51.05 \pm 0.56\%}$ \\
		\midrule
		MetaSGD~\cite{li2017metasgd} & 
		$34.52 \pm 0.63\%$ & $49.22 \pm 0.58\%$ \\
		MetaSGD w/ Dropout~\cite{srivastava2014dropout} &
		$35.01 \pm 0.54\%$ & $52.35 \pm 0.58\%$ \\
		MetaSGD w/ DropGrad &
		$\mathbf{36.77 \pm 0.72\%}$ & $\mathbf{55.13 \pm 0.72\%}$ \\
		\midrule
		MetaSGD* & $43.98 \pm 0.77\%$ & $57.95 \pm 0.81\%$ \\
		MetaSGD* w/ DropGrad & $\mathbf{45.33 \pm 0.81\%}$ & $\mathbf{59.94 \pm 0.82\%}$ \\
		\midrule
    	MAML++~\cite{antoniou2018maml++} &
    	$40.73 \pm 0.49\%$ & $60.57 \pm 0.49\%$ \\
    	MAML++ w/ Dropout~\cite{srivastava2014dropout} &
    	$41.75 \pm 0.49$\% & $61.48 \pm 0.49\%$ \\
		MAML++ w/ DropGrad &
    	$\mathbf{44.27 \pm 0.50\%}$ & $\mathbf{63.79 \pm 0.48\%}$ \\
		\bottomrule
	\end{tabular}
	\label{tab:cross}
	\vspace{-4mm}
\end{table}
%

% ------------------------------------------------------------ %
\begin{figure*}[t]
    \centering
    \includegraphics[width=\textwidth]{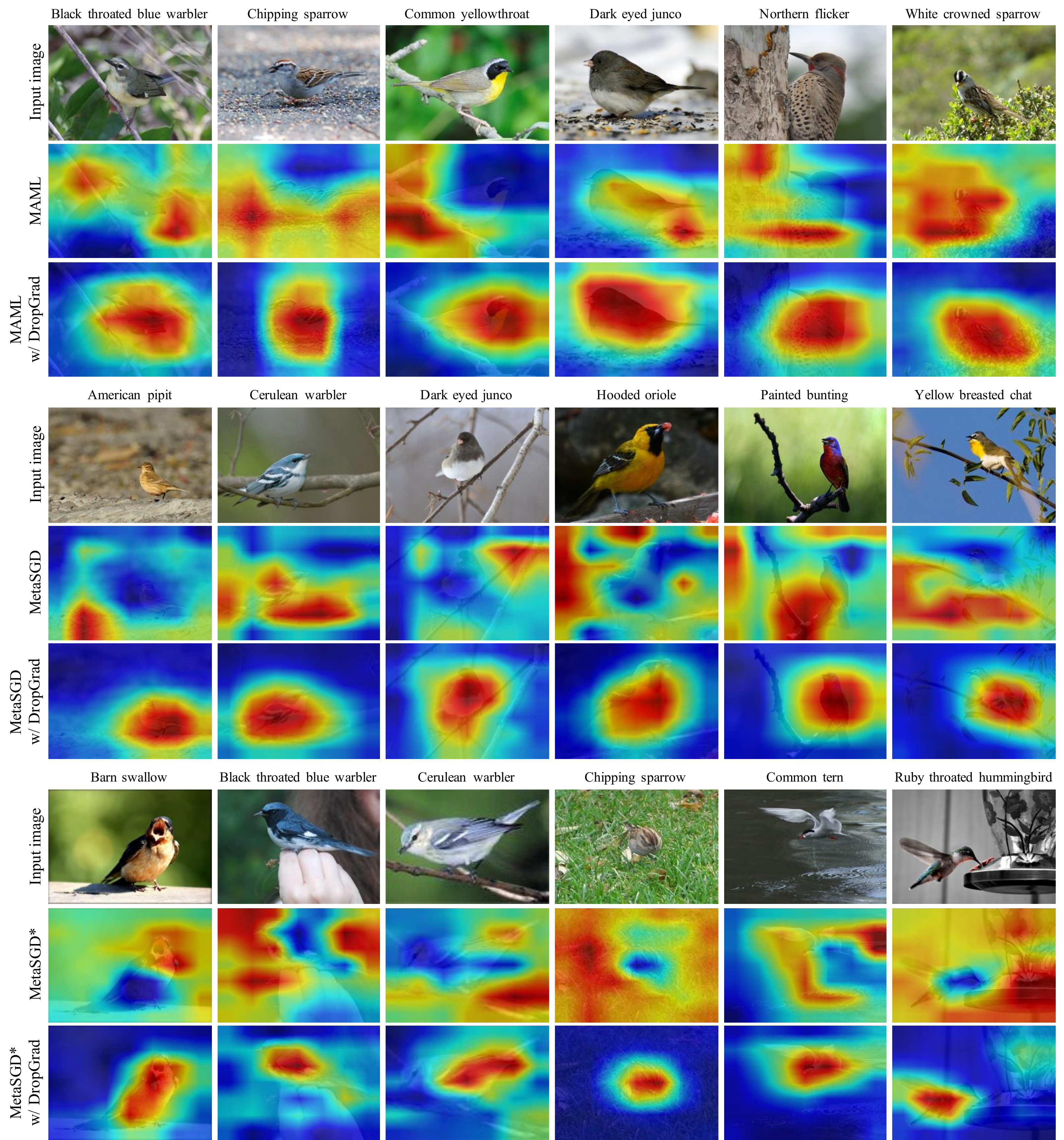}
    %\vspace{-2mm}
    \caption{\textbf{Class activation maps (CAMs) for cross-domain 5-shot classification.} The mini-ImageNet and CUB datasets are used for the meta-training and meta-testing steps, respectively. Models trained with the proposed DropGrad (the third row for each example) focus more on the objects than the original models (the second row for each example).}
    \label{fig:cam}
    \vspace{-4mm}
\end{figure*}
\subsection{Cross-Domain Few-Shot Classification}
%\vspace{-2mm}

%\paragraph{Cross-Domain Scenario}
%
To further evaluate how the proposed DropGrad method improves the generalization ability of gradient-based meta-learning models, we conduct a cross-domain experiment, in which the meta-testing set is from an \textit{unseen} domain.
%
% We further investigate the impact of the proposed Gaussian DropGrad on generalizing gradient-based meta-learning models to \textit{unseen} domain.
%
We use the cross-domain scenario introduced by Chen~\etal~\cite{chen2019closerfewshot}, where the meta-training step is performed on the mini-ImageNet~\cite{vinyals2016matching} dataset while the meta-testing evaluation is conducted on the CUB~\cite{hilliard2018few} dataset.
Note that, different from Chen~\etal~\cite{chen2019closerfewshot} who select the model according to the validation performance on the CUB dataset, we pick the model via the validation performance on the mini-ImageNet dataset for evaluation.
The reason is that we target at analyzing the generalization ability to the unseen domain, and thus we do not utilize any information provided from the CUB dataset.

\tabref{cross} shows the results using the Gaussian DropGrad method. 
Since the domain shift in the cross-domain scenario is larger than that in the intra-domain case (\ie, both training and testing tasks are sampled from the mini-ImageNet dataset), the performance gains of applying the proposed DropGrad method reported in \tabref{cross} are more significant than those in \tabref{mini}. 
The results demonstrate that the DropGrad scheme is able to effectively regularize the gradients and transfer them for learning new tasks in an unseen domain.

To further understand the improvement by the proposed method under the cross-domain setting, we visualize the class activation maps (CAMs)~\cite{zhou2016learning} of the images in the unseen domain~(CUB).
More specifically, during the testing time, we adapt the learner model $f_\theta$ with the support set $D^s$.
We then compute the class activation maps of the data in the query set $D^q$ from the last convolutional layer of the updated learner model $f_\theta'$.
\figref{cam} demonstrates the results of the MAML, MetaSGD, and MetaSGD* approaches.
The models trained with the proposed regularization method show the activation on more discriminative regions.
This suggests that the proposed regularization improves the generalization ability of gradient-based schemes, and thus enables these methods to adapt to the novel tasks sampled from the unseen domain.

\vspace{-2mm}
\paragraph{Comparison with the Existing Dropout Approach.}
We also compare the proposed DropGrad approach with existing Dropout~\cite{srivastava2014dropout} method under the cross-domain setting.
We apply the existing Dropout scheme on the network activations in both $f_\theta$ and $f_\theta'$.
As suggested by Ghiasi~\etal~\cite{ghiasi2018dropblock}, we use the dropout rate of $0.3$ for the Dropout method.
As the results shown in~\tabref{cross}, the proposed DropGrad method performs favorably against the Dropout approach.
The larger performance gain from the DropGrad approach validates effectiveness of imposing uncertainty on the inner-loop gradient for the gradient-based meta-learning framework.
On the other hand, since applying the conventional Dropout causes the inconsistent randomnesses between two different sets of parameters $f_\theta$ and $f_\theta'$, which is less effective compared to the proposed scheme.

%\begin{figure*}[t]
%    \centering
%    \includegraphics[width=\textwidth]{figure/crest.png}
%    \includegraphics[width=\textwidth]{figure/sdnet.png}
%    \caption{\textbf{Precision and success plots on OTB2015 dataset.}}
%    \label{fig:meta_tracker}
%\end{figure*}

\begin{table}[]
	\centering
	\footnotesize
	\caption{\textbf{Precision and success rate on the OTB2015 dataset.} The DropGrad method can be applied to visual tracking and improve the tracking performance. }
    \begin{tabular}{l cc}
        \toprule
        Model                           & Precision         & Success rate      \\ \midrule
        MetaCREST~\cite{park2018meta}   & $0.7994$          & $0.6029$          \\
        MetaCREST w/ DropGrad & $\mathbf{0.8172}$ & $\mathbf{0.6145}$ \\ \midrule
        MetaSDNet~\cite{park2018meta}   & $0.8673$          & $0.6434$          \\
        MetaSDNet w/ DropGrad & $\mathbf{0.8746}$ & $\mathbf{0.6520}$ \\ \bottomrule
    \end{tabular}
	\label{tab:meta_tracker}
	\vspace{-4mm}
\end{table}

% ------------------------------------------------------------ %
\begin{figure*}[t]
    \centering
    \includegraphics[width=\textwidth]{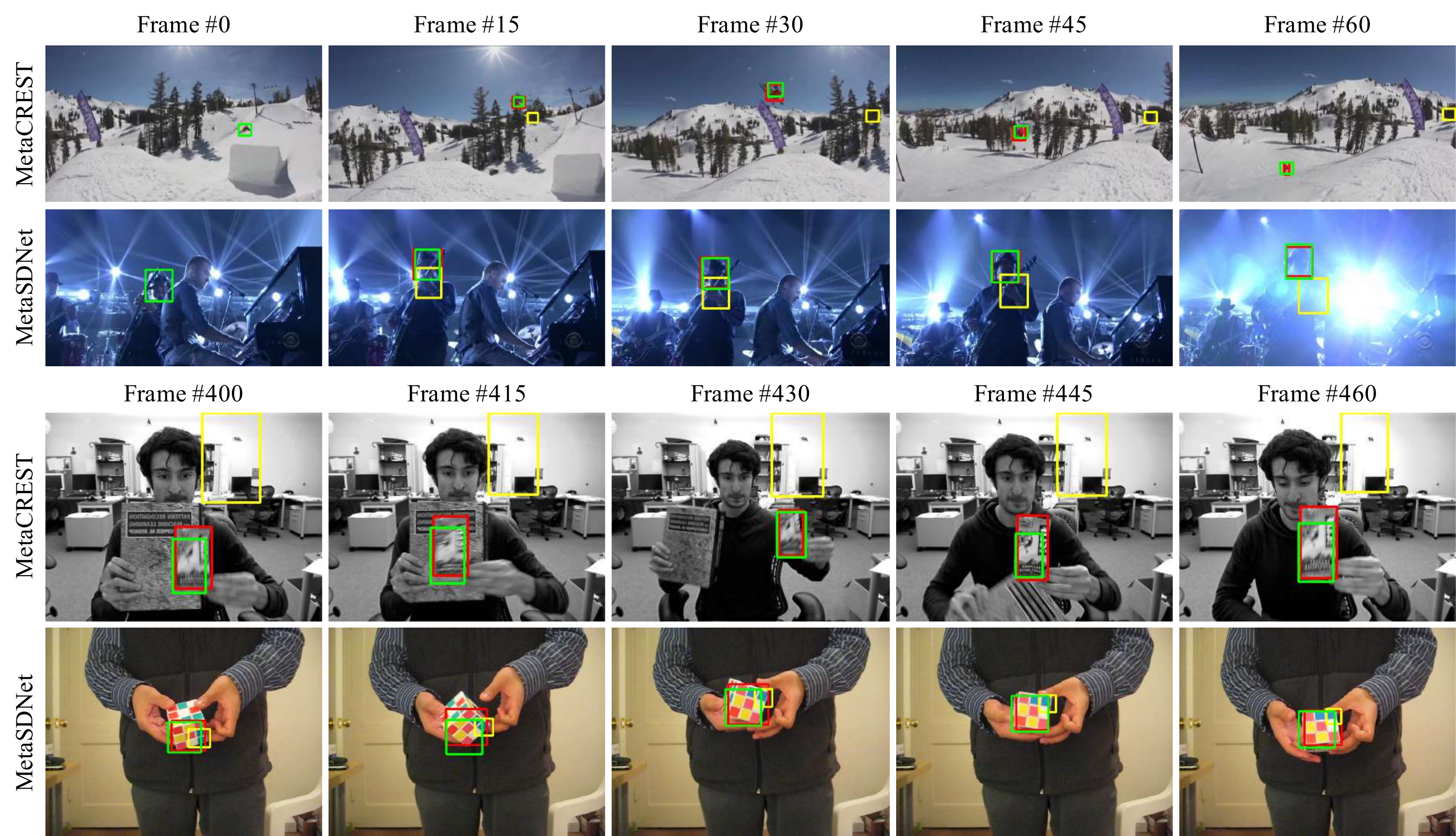}
    %\vspace{-2mm}
    \caption{\textbf{Qualitative results of object online tracking on the OTB2015 dataset.} Red boxes are the ground-truth, yellow boxes represent the original results, and green boxes stand for the results where the DropGrad method is applied. Models trained with the proposed DropGrad scheme are able to track objects more accurately.}
    \label{fig:meta_tracker}
    \vspace{-4mm}
\end{figure*}
\subsection{Online Object Tracking}
%\vspace{-2mm}
Visual object tracking targets at localizing one particular object in a video sequence given the bounding box annotation in the first frame.
To adapt the model to the subsequent frames, one approach is to apply online adaptation during tracking.
The Meta-Tracker~\cite{park2018meta} method uses meta-learning to improve two state-of-the-art online trackers, including the correlation-based CREST~\cite{song2017crest} and the detection-based MDNet~\cite{nam2016mdnet}, which are denoted as MetaCREST and MetaSDNet.
Based on the error signals from future frames, the Meta-Tracker updates the model during offline meta-training, and obtains a robust initial network that generalizes well over future frames.
We apply the proposed DropGrad method to train the MetaCREST and MetaSDNet models with evaluation on the OTB2015~\cite{wu2015otb} dataset.

\vspace{-2mm}
\paragraph{Implementation Details.}
We train the models using the original source code.\footnote{https://github.com/silverbottlep/meta\_trackers}
%(https://github.com/silverbottlep/meta\_trackers).
%\footnote{https://github.com/silverbottlep/meta\_trackers}
%
For meta-training, we use a subset of a large-scale video detection dataset~\cite{russakovsky2015ILSVRC}, and the $58$ sequences from the VOT2013~\cite{kristan2013vot}, VOT2014~\cite{kristan2014vot} and VOT2015~\cite{kristan2015vot} datasets, excluding the sequences in the OTB2015 database, based on the same settings in the Meta-Tracker~\cite{park2018meta}.
We apply the Gaussian DropGrad method with the dropout rate of $0.2$.
We use the default hyper-parameter settings and evaluate the performance with the models at the last training iteration.

\vspace{-2mm}
\paragraph{Object Tracking Results.}
The results of online object tracking on the OTB2015 dataset are presented in \tabref{meta_tracker}.
The one-pass evaluation (OPE) protocol without restarts at failures is used in the experiments. 
We measure the precision and success rate based on the center location error and the bounding-box overlap ratio, respectively.
The precision is calculated with a threshold $20$, and the success rate is the averaged value with the threshold ranging from $0$ to $1$ with a step of $0.05$.
We show that applying the proposed DropGrad method consistently improves the performance in precision and success rate on both MetaCREST and MetaSDNet trackers.

We present sample results of object online tracking in \figref{meta_tracker}.
We apply the proposed DropGrad method on the MetaCREST and MetaSDNet methods and evaluate these models on the OTB2015 dataset.
Compared with the original MetaCREST and MetaSDNet, models trained with the DropGrad method track objects more accurately.

\subsection{Few-Shot Viewpoint Estimation}
Viewpoint estimation aims to estimate the viewpoint~(\ie, 3D rotation), denoted as $R\in{\mathrm{SO}(3)}$, between the camera and the object of a specific category in the image.
Given a few examples~(\ie, $10$ images in this work) of a novel category with viewpoint annotations, few-shot viewpoint estimation attempts to predict the viewpoint of arbitrary objects of the same category.
In this problem, the support set $D^s$ contains few images $\mathbf{x}^s$ of a new class and the corresponding viewpoint annotations $\mathbf{y}^s$.
We conduct experiments on the ObjectNet3D~\cite{xiang2016objectnet3d} dataset,
a viewpoint estimation benchmark dataset which contains $100$ categories.
Using the same evaluation protocol in~\cite{tseng2019metaview}, we extract $76$ and $17$ categories for training and testing, respectively.

\vspace{-2mm}
\paragraph{Implementation Details.}
We apply the proposed DropGrad on the MetaView~\cite{tseng2019metaview} method, which is a meta-Siamese viewpoint estimator that applies gradient-based adaptation for novel categories.
We obtain the source code from the authors, and keep all the default setting for training.
We apply the Gaussian DropGrad scheme with the dropout rate of $0.1$.
Since there is no validation set available, we pick the model trained in the last epoch for evaluation.

\vspace{-2mm}
\paragraph{Viewpoint Estimation Results.}
We show the viewpoint estimation results in \tabref{viewpoint}.
The evaluation metrics include Acc30 and MedErr, which represent the percentage of viewpoints with rotation error under $30^{\circ}$ and the median rotation error, respectively.
The overall performance is improved by applying the proposed DropGrad method to the MetaView model during training.

% ------------------------------------------------------------ %
\begin{table}[t]
	\centering
	\footnotesize
	%\small
	%\setlength\tabcolsep{2.5pt}
	\caption{\textbf{Viewpoint estimation results.} The DropGrad method can be applied to few-shot viewpoint estimation frameworks to mitigate the overfitting problem.}
	\begin{tabular}{l cc} 
	    \toprule
		Model & Acc30 ($\uparrow$) & MedErr ($\downarrow$) \\
		\midrule
		MetaView~\cite{tseng2019metaview} & 
		$45.00 \pm 0.45\%$ & $33.60 \pm 0.94^{\circ}$ \\
		MetaView w/ DropGrad & 
		$\mathbf{46.16 \pm 0.55\%}$ &  $\mathbf{33.10 \pm 0.82^{\circ}}$\\
		\bottomrule 
	\end{tabular}
	\label{tab:viewpoint}
	\vspace{-5mm}
\end{table}
% ------------------------------------------------------------ %
\vspace{-2mm}
\section{Conclusions}
\label{sec:conclusions}

% Briefly summarize what we did in this work.
In this work, we propose a simple yet effective gradient dropout approach for regularizing the training of gradient-based meta-learning frameworks.
The core idea is to impose uncertainty by augmenting the gradient in the adaptation step during meta-training.
We propose two forms of noise regularization terms, including the Bernoulli and Gaussian distributions, and demonstrate that the proposed DropGrad improves the model performance in three learning tasks.
In addition, extensive analysis and studies are provided to further understand the benefit of our method.
One study on cross-domain few-shot classification is also conducted to show that the DropGrad method is able to mitigate the overfitting issue under a larger domain gap.
% Disclose all limitations

% Describe how potential future work can address these limitations and lead to interesting and ground breaking stuff.

{\small
\bibliographystyle{ieee_fullname}
\bibliography{main}
}
\clearpage

\appendix
\section{Appendix}
In this appendix, we first supplement the implementation details. We then present additional experimental results of reinforcement learning and cross-domain few-shot classification.
Finally, we compare the proposed DropGrad regularization algorithm with the simulated annealing methods.

\section{Supplementary Implementation Details}
\paragraph{Few-Shot Classification.}
We use the implementation from \cite{chen2019closerfewshot} to train and evaluate MAML~\cite{finn2017maml} on few-shot classification tasks.\footnote{https://github.com/wyharveychen/CloserLookFewShot}
In the meanwhile, we modify the same implementation for MetaSGD~\cite{li2017metasgd} by ourselves.
We verify our implementation by evaluating the MetaSGD model using Conv4, which is the same backbone network adopted in the original paper.
The $5$-way $5$-shot classification results on the mini-ImageNet dataset~\cite{ravi2017metalstm} reported by our implementation and the original paper are $65.31 \pm 0.66\%$ and $64.03 \pm 0.94\%$, respectively.

To train both the MAML and MetaSGD models, we keep the default settings in the original implementation by \cite{chen2019closerfewshot}.
We apply the Adam~\cite{kinga2015method} optimizer with the learning rate of $0.001$.
The mini-batch size is set to be $4$.
We train the model with $400$ epochs and do not apply the learning rate decay strategy.

\paragraph{Online Object Tracking.}
We conduct experiments of online object tracking based on the PyTorch implementation by \cite{park2018meta}.
For MetaSDNet, the first three convolutional layers of VGG-16 are used as the feature extractor.
During meta-training, the last three fully-connected layers are randomly initialized. We only update the last three fully-connected layers in the first 5,000 iterations, and then train the entire network for the remaining iterations.
We adopt the Adam optimizer~\cite{kinga2015method} with an initial learning rate of $10^{-4}$, and decrease the learning rate to $10^{-5}$ after $10,000$ iterations. In total, we train the network for $15,000$ iterations.
For MetaCREST, we use the Adam optimizer with a learning rate of $10^{-6}$, and train the model for $10,000$ iterations.

\section{Reinforcement Learning}
We adopt the few-shot reinforcement learning~(RL) setting as in~\cite{finn2017maml}, which aims to make the system adapt to new experiences and learn the corresponding policy quickly with limited prior experience (\ie, trajectories).
In this setting, the support set $D^s$ contains few trajectories and the corresponding rewards, while the query set $D^q$ is formed by a set of new trajectories sampled from the running policy.
We conduct the experiment with the locomotion tasks simulated by the MuJoCo~\cite{todorov2012mujoco} simulator.
Two environments are considered in this experiment: HalfCheetah robot and Ant robot with forward/backward movement, \ie, HalfCheetah-Dir and Ant-Dir.

\paragraph{Implementation Details.}
We adopt the MAML-TPRO~\cite{finn2017maml} framework as the baseline method.
Since the rewards are usually not differentiable, policy gradients are calculated for adapting the RL models to new experiences in both inner- and outer-loop optimization.
For applying the proposed DropGrad scheme in the RL framework, we augment the policy gradients calculated according to rewards in the support set during the inner-loop optimization.
%
% The proposed DropGrad scheme is applied to augment the policy gradients calculated according to rewards in the support set in the training stage.
%
We use a public PyTorch implementation with the default hyper-parameter settings in the experiments.\footnote{https://github.com/tristandeleu/pytorch-maml-rl}

\begin{figure}[t]
    \centering
    \includegraphics[width=\linewidth]{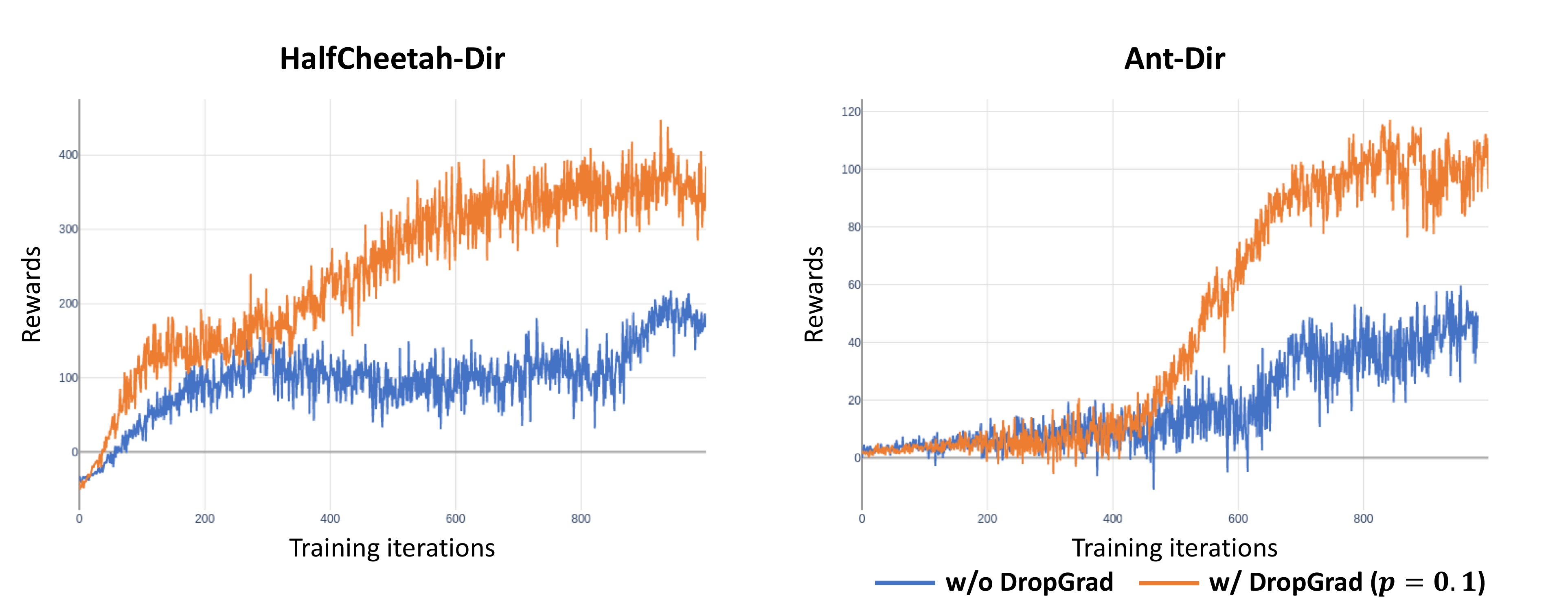}
    \caption{\textbf{Few-shot reinforcement learning results.} Two settings, HalfCheetah-Dir (\textit{left}) and Ant-Dir (\textit{right}), are considered in our experiments using the MAML-TPRO framework. We show the reward curves after the model is updated with the few trajectories and both rewards converge favorably against the original training.}
    \label{fig:rl}
    \vspace{-4mm}
\end{figure}

\paragraph{Reinforcement Learning Results.}
In \figref{rl}, we present the rewards after the model is optimized with the few trajectories, \ie, in each iteration we perform a one-step policy gradient update for the inner-loop optimization.
In both environments, the training process with the proposed DropGrad regularization method converges to favorable rewards compared to the original training without the proposed regularization.
This improvement could be attributed to the uncertainty on gradients that provides a better exploration of the policy.

\section{Cross-Domain Few-Shot Classification}
\begin{figure*}[t]
    \centering
    \includegraphics[width=\textwidth]{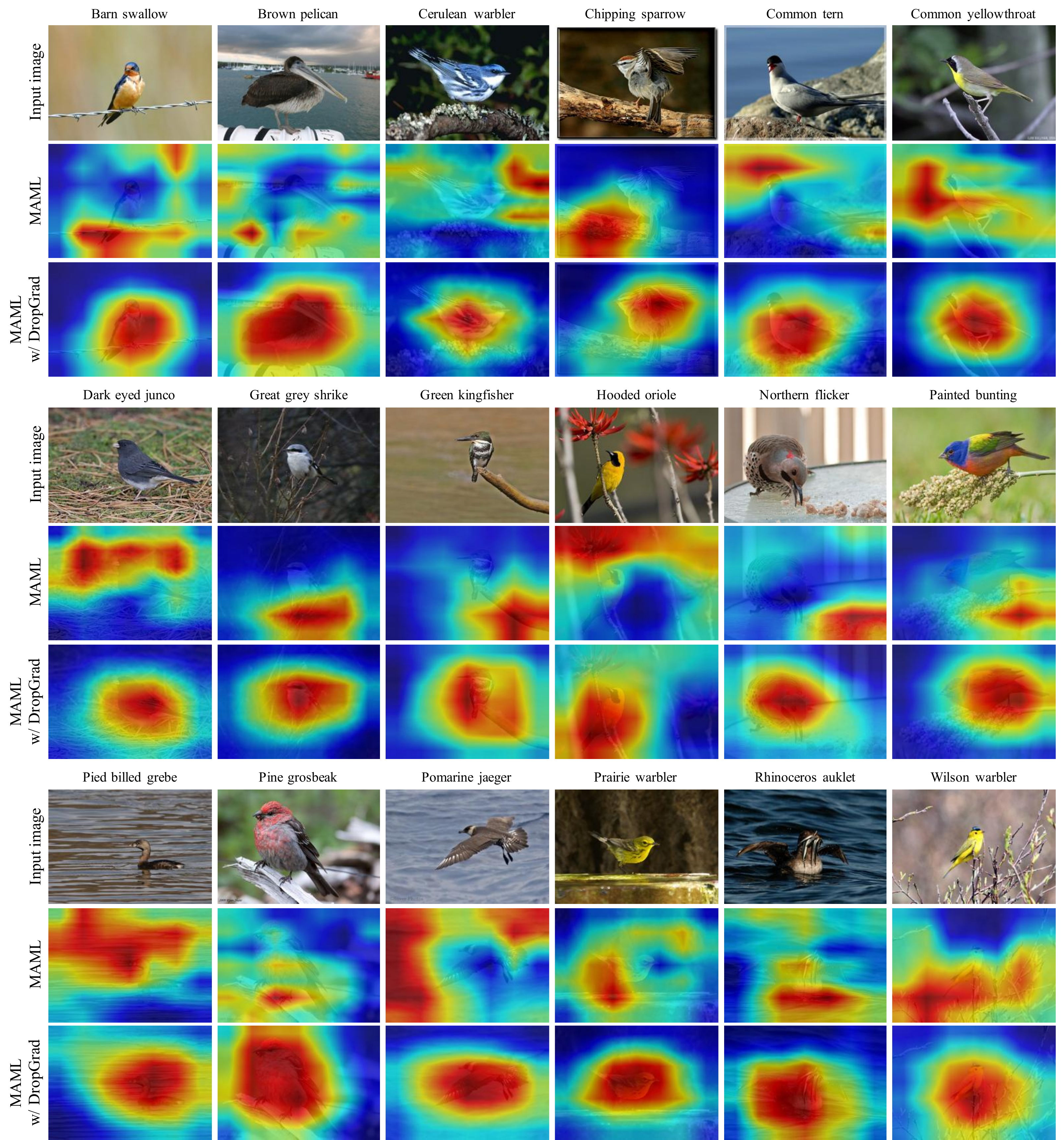}
    %\vspace{-2mm}
    \caption{\textbf{Class activation maps (CAMs) for cross-domain 5-shot classification.} The mini-ImageNet and CUB datasets are used for the meta-training and meta-testing steps, respectively. Models trained with the proposed DropGrad (the third row for each example) focus more on the objects than the original models (the second row for each example).}
    \label{fig:cam_maml}
    %\vspace{-4mm}
\end{figure*}

\begin{figure*}[t]
    \centering
    \includegraphics[width=\textwidth]{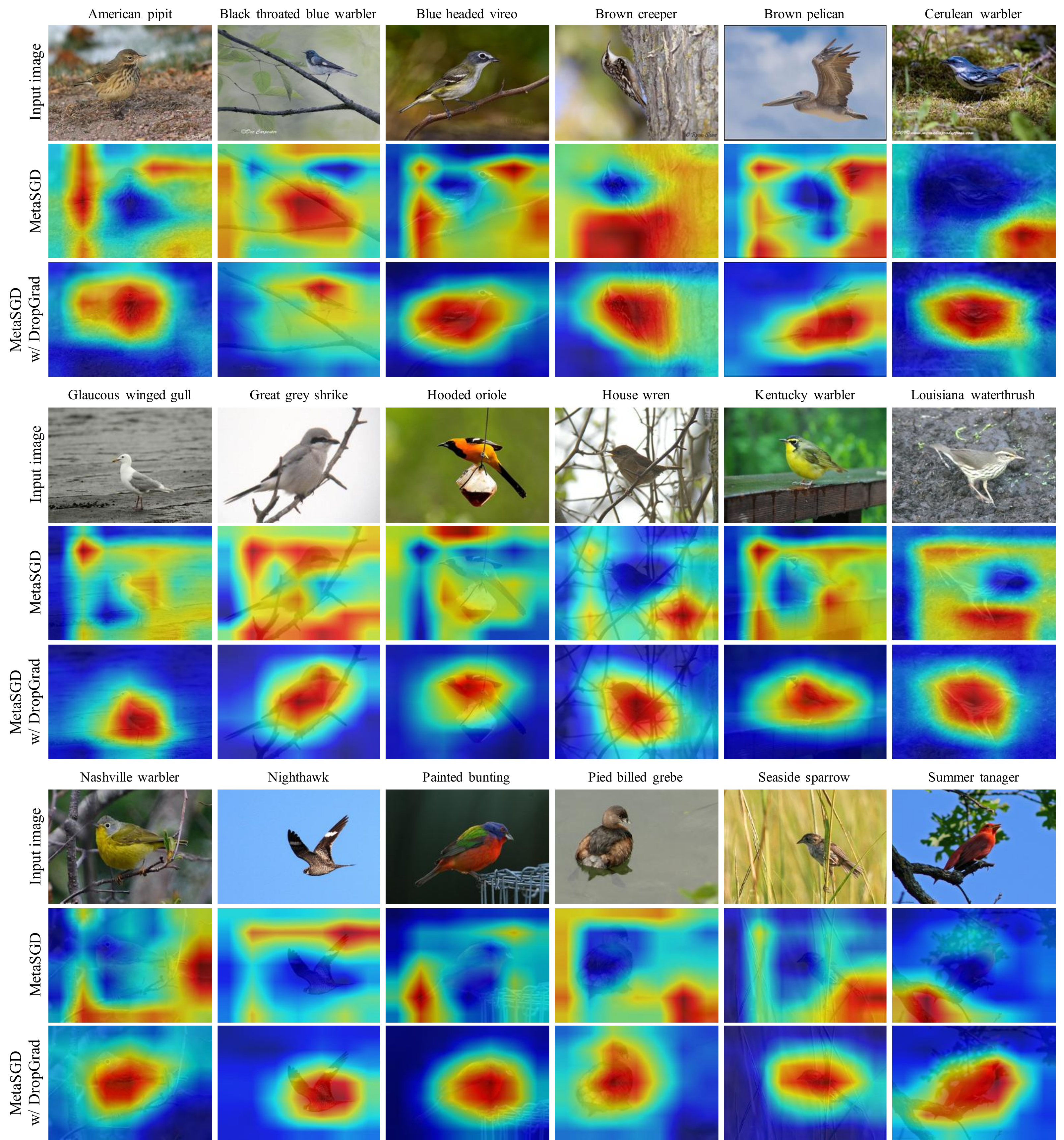}
    %\vspace{-2mm}
    \caption{\textbf{Class activation maps (CAMs) for cross-domain 5-shot classification.} The mini-ImageNet and CUB datasets are used for the meta-training and meta-testing steps, respectively. Models trained with the proposed DropGrad (the third row for each example) focus more on the objects than the original models (the second row for each example).}
    \label{fig:cam_metasgd}
    %\vspace{-4mm}
\end{figure*}

\begin{figure*}[t]
    \centering
    \includegraphics[width=\textwidth]{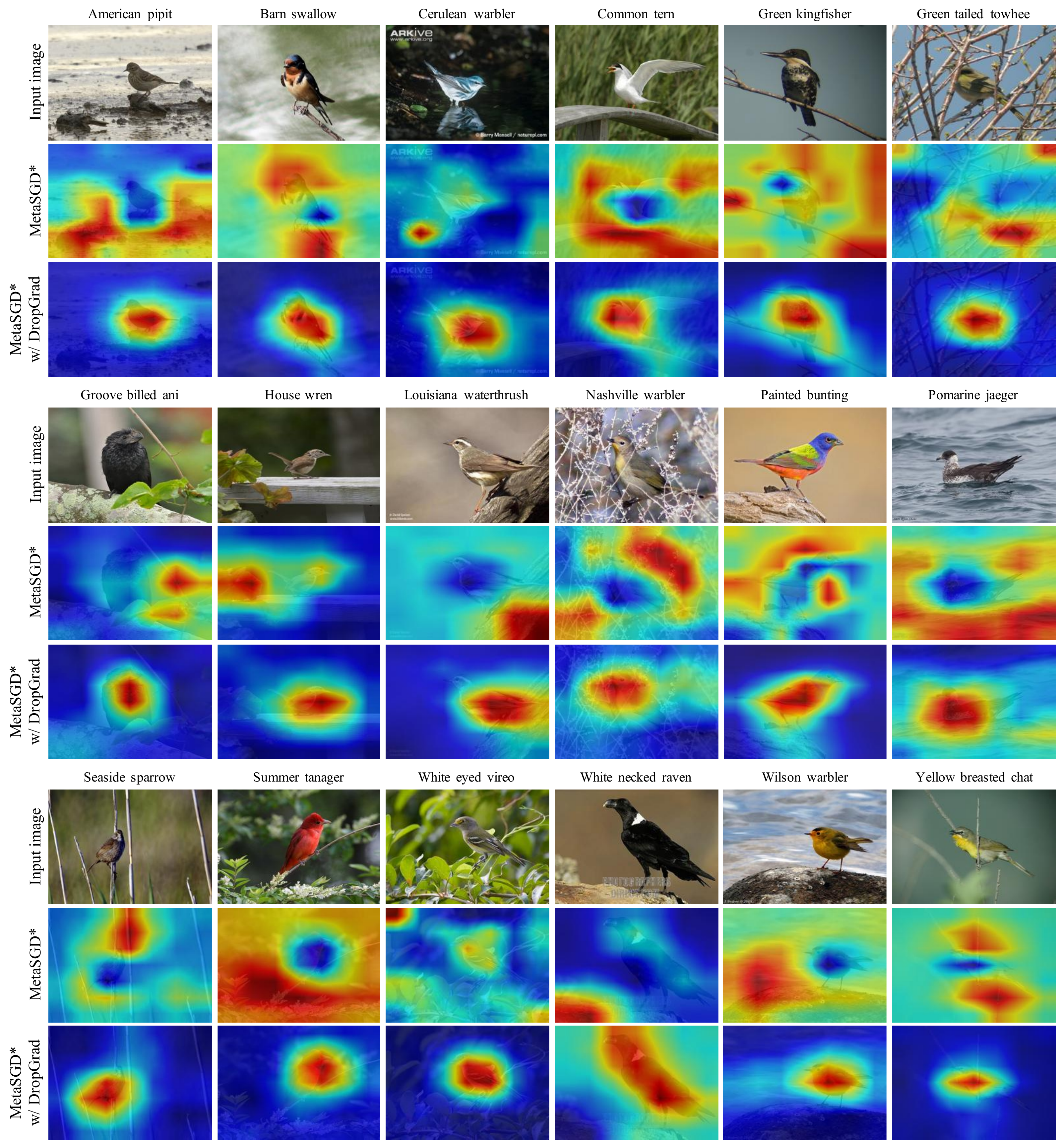}
    %\vspace{-2mm}
    \caption{\textbf{Class activation maps (CAMs) for cross-domain 5-shot classification.} The mini-ImageNet and CUB datasets are used for the meta-training and meta-testing steps, respectively. Models trained with the proposed DropGrad (the third row for each example) focus more on the objects than the original models (the second row for each example).}
    \label{fig:cam_metasgd_warmup}
    %\vspace{-4mm}
\end{figure*}

In \figref{cam_maml}, \figref{cam_metasgd} and \figref{cam_metasgd_warmup}, we provide more results of the class activation maps (CAMs)~\cite{zhou2016learning} for the cross-domain few-shot classification task.
The meta-training and meta-testing steps are conducted on the mini-ImageNet and CUB datasets, respectively.
We apply the DropGrad scheme on the MAML, MetaSGD, and MetaSGD* approaches.
The results show that models trained with the proposed DropGrad regularization focus on more discriminative regions.

% ------------------------------------------------------------ %
% ------------------------------------------------------------ %
% ------------------------------------------------------------ %
% ------------------------------------------------------------ %

%\section{Object Online Tracking}
%We present more results of object online tracking in \figref{meta_tracker}.
%
%We apply the proposed DropGrad method on the MetaCREST and MetaSDNet methods and evaluate these models on the OTB2015 dataset.
%
%Compared with the original MetaCREST and MetaSDNet, models trained with the DropGrad method track objects more accurately.

\section{Comparison to Simulated Annealing}
The proposed DropGrad algorithm is also related to simulated annealing~(SA)~\cite{kirkpatrick1983optimization}.
While conceptually similar to a certain extent, the goals and formulations are significantly different. 
SA modulates gradients by exploring uncertain solutions to escape from the local minimum during the training stage.
On the other hand, our DropGrad method drops the \textit{inner} gradient to introduce uncertainty in the forward pass of the gradient-based meta-learning framework.

\end{document}